\documentclass{IEEEtran}
\usepackage{cite}
\usepackage{amsmath,amssymb,amsfonts}
\usepackage{algorithmic}
\usepackage{graphicx}
\usepackage{textcomp}
\usepackage{math}
\usepackage{booktabs, multirow}  
\def\BibTeX{{\rm B\kern-.05em{\sc i\kern-.025em b}\kern-.08em
    T\kern-.1667em\lower.7ex\hbox{E}\kern-.125emX}}
\begin{document}
\title{Adversarial Multi-scale Feature Learning for Person Re-identification}
\author{Xinglu Wang 
	  \thanks{Xinglu Wang is with College of Information Science  Electronic Engineering, China. (Email: \texttt{xingluwang@zju.edu.cn})  
	  } 
}
	
\maketitle

\begin{abstract}
Person Re-identification (Person ReID) is an important topic in intelligent surveillance and computer vision. It aims to accurately measure visual similarities between person images for determining whether two images correspond to the same person. 
The key to accurately measure visual similarities is learning  discriminative features, which not only captures clues from different spatial scales, but also jointly inferences on multiple scales, with the ability to determine reliability and ID-relativity of each clue.   
To achieve these goals, 
we propose to improve Person ReID system performance from two perspective: 
\textbf{1).} Multi-scale feature learning (MSFL), which consists of Cross-scale information propagation (CSIP) and Multi-scale feature fusion (MSFF), to dynamically fuse features cross different scales.
\textbf{2).} Multi-scale gradient regularizor (MSGR), to emphasize ID-related factors and ignore irrelevant factors in an adversarial manner. 
Combining MSFL and MSGR, our method achieves the state-of-the-art performance on four commonly used person-ReID datasets with neglectable test-time computation overhead.  
\end{abstract}


\section{Introduction} \label{sec:intro}

Given a probe image, Person ReID aims to identify the person of interest from a large gallery image database 
collected from different cameras. 
Person ReID is conducted by estimating the visual similarities between persons. 
The gallery images are ranked in the descending order of the similarities as re-identification results. 
Such a task has extensive applications in intelligent surveillance. For instance,
it can be used to search criminal suspects or missing persons from a large surveillance camera network 
efficiently and effectively. 

\begin{figure}
	\centering
	\includegraphics[width=0.99\linewidth]{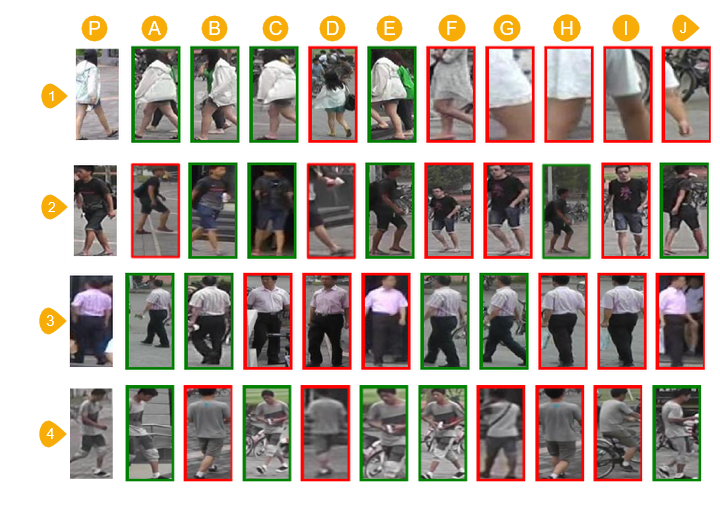}
	\caption{Several retrieval results for a converged ReID model.
		Each row represents one visual case. First column (P) is the probe image. A -- J columns show the retrieved  gallery images, where the green boxes denote true positives and the red boxes denote false negatives. 
		From these visual cases, we observe various challenges in Person ReID task that can be summarized into two aspects: 
		\textbf{1).} scale variance of clues. 
		\textbf{2).} reliability of each clue.
		To name the specific examples, 
		\textbf{1).} detection misalignment (Row 1, Column D) causes the large scale and shift variance, 
		\textbf{2).} False alarm of a detector (Row 2, Column F) produces partial human body part, which serves as a distractor in Test-set; 
		\textbf{3).} Extremely confusing identity pairs (Row 2, Column A) requires scrutiny on multi-scale clues, including not only the global color attribute, but also  tiny details like bags and shoes.
		\textbf{4).} Global appearance like color and body structure helps, but is influenced by camera parameters and human pose,\eg, the color distortion caused by camera settings shown in Row 3. 
		\textbf{5).} Facial attributes like wearing glass are useful in Row 2, however, face gets blur or partially occluded in Row 3, whose importance should be reduced. 
		\textbf{6).} Local parts  of different scales are not certainly reliable, \eg, misalignment causes the missing of shoes (Row 2 ,Column C), 
		suddenly appear of a bicycle in the background (Row 4, Column E),   the appearance change of bag caused by different poses (Row 2, Column E). 
	}
	\label{fig:challenge}
\end{figure}

As illustrated in Fig.~\ref{fig:challenge}, 
person ReID faces many challenges, which are summarized as two aspects: 
\textbf{1).} The scale of clues to distinguish persons changes dramatically. 
\textbf{2).} None of the clues is absolutely reliable to distinguish persons. 
To give the specific examples, 
\textbf{1).} Detection misalignment (Row 1, Column D) causes the large scale and shift variance, 
\textbf{2).} False alarm of a detector (Row 2, Column F) produces partial human body part, which serves as a distractor in testset; 
\textbf{3).} The existence of extremely similar persons (Row 2, Column A)  needs a careful scrutiny of multi-scale clues, including not only the global color attribute, but also  tiny details like bags and shoes, and intelligently inference upon them.
\textbf{4).} Global appearance like color and body structure helps, but is influenced by camera parameters and human pose, \eg, the color distortion caused by camera settings in Row 3. 
\textbf{5).} Facial attributes like wearing glass are useful in Row 2, however, face gets blur or partially occluded in Row 3, and importance of face in this circumstance should be reduced. 
\textbf{6).} Local parts of different scales are not certainly reliable, \eg, misalignment causes the missing of shoes (Row 2 ,Column C), 
the bicycle emerging in the background (Row 4, Column E), the appearance change of bag caused by different poses (Row 2, Column E). 

We argue that all the challenge would be handled when the model is: 
\textbf{1).} Capable of discovering global-scale and multiple local-scale clues for distinguishing person from person. 
\textbf{2).} Able to evaluate the reliability and ID-relativity of each clue, and finally combine the clues into a discriminative feature.  

To achieve the goals mentioned above, we propose two improvement: 
\textbf{1).} Multi-scale feature learning (MSFL), with a model structure possessing the potential to utilize different scales and their combination, and intelligently fuse various scales. 
\textbf{2).} To fully exploit and realize the potential, we guide the model with Multi-scale Gradient Regularizer (MSGR), which implicitly consider all possible perturbation in the beginning and gradually emphasize adversarial perturbation as model approaching converge. In this way, the model converges to a better optimum, learns to ignore irrelevant factors and focuses on the combination of informative ID-related factors.

\section{Related Work}

\subsection{Pyramid Structure} 

Due to the success of deep learning, CNNs have emerged as general purpose feature extractors for a wide range of visual recognition tasks. 
However, the features from single convolution layer are insufficient for many tasks that need multi-scale clues to inference.  
The importance of multi-scale feature learning and the design of pyramid network structure has been recognized recently 
and some recent works \cite{cai2017higher,hariharan2015hypercolumns,long2015fully,xie2015holistically,you2018structurally} attempt to investigate the effectiveness of exploiting feature from different convolution layers within a CNN.  
For example, Hariharan et al. \cite{hariharan2015hypercolumns} considered the feature maps from all convolution layers, allowing finer grained resolution for localization tasks. 
Long et al. \cite{long2015fully} combined the finer-level and higher-level semantic feature from different convolution layers for better segmentation. 
Xie et al. \cite{xie2015holistically} proposed a holistically-nested framework where the side outputs are added after lower convolution layers to provide deep supervision for edge detection. 
Cai et al. \cite{cai2017higher} concatenated the activation maps from multiple convolution layers to model the interaction of part features for fine-grained recognition. 

However, simply concatenating multi-scale features may fails to capture the importance, reliability, semantically abstract level and ID-relativity of features in different scales, yields a inferior model. 
Admittedly, the concatenation of hierarchical features incorporates information of different spatial resolutions. However, it also introduces large semantic gaps caused by different depths. 
The high-resolution low-semantic maps may harm their representational capacity for overall task. 
Due to this reason, the Single Shot Detector (SSD) \cite{liu2016ssd} foregoes reusing low-level features, instead builds the pyramid starting from high up in the network (e.g., conv4\_3 of VGG nets \cite{simonyan2014very}) and then by adding several new layers. However, the SSD-style pyramid misses the opportunity to reuse the higher-resolution maps of the feature hierarchy. 

To create a feature pyramid that has strong semantics at all scales, many novel neural network architectures have been proposed, including U-Net \cite{ronneberger2015u} and Sharp-Mask \cite{pinheiro2016learning} for segmentation, Recombinator networks \cite{honari2016recombinator} for face detection, and Stacked Hourglass networks \cite{newell2016stacked} for keypoint estimation. Ghiasi et al. \cite{ghiasi2016laplacian} present a Laplacian pyramid presentation for FCNs to progressively refine segmentation. 
These methods adopt architectures with pyramidal shapes, and exploit lateral/skip connections that associate low-level feature maps with high-semantic features. Different to their works, we further improves the representation power of feature according to its pyramid level.  

%

\subsection{Adversarial Learning} 

The idea of adversarial learning comes from adversarial training \cite{goodfellow2014explaining}, 
where  a mixture of normal and adversarially-generated examples are applied in the training process, in the hope to increase the robustness against adversarial examples. However, as suggested by \cite{ross2018improving}, inject adversarial noise  in a anisotropic and brute-force way harms the model performance 
on the original validation set,
since adversarially-generated examples are not naturally interpretable images. 

Adversarial learning originated from regularizing techniques that reduce overfitting, where an implicit or explicit player turns against the optimization goal. 
For example, Bengio et al.\cite{bengio2013estimating} and Gulcehre et al.\cite{gulcehre2016noisy} add noise in the ReLU and Sigmoid activation functions respectively.
Szegedy et al. \cite{szegedy2016rethinking} propose label-smoothing regularization technique to minimize distance between the model distribution and uniform distribution.  

Metric learning community start to explore this advanced learning paradigm recently. For example, 
Deep adversarial metric learning \cite{duan2018deep} simultaneously learns to generate synthetic hard negatives from the observed easy negative samples and discriminate the feature embedding in an adversarial manner. 
Adversarial metric learning \cite{Chen2018adv} simultaneously train the hard negative generator and feature embedding in an adversarial manner. 
Energy confusion regularization \cite{chen2019energy} seeks to confuse the learned model by enlarging intra-variance of all positive samples. 
All of them design a adversarial target, but optimized in one-step and alternative updating fashion. 
Different from existing work, we design a differentiable gradient regularizer that implicitly generate multiply adversarial perturbation at the same time and jointly optimized with original target.

\section{Proposed Method} 

Generally, representation learning tasks aims at learning a compact representation,
from which the downstream tasks benefit. 
As a special case of representation learning task, the training pipeline of  Person ReID is also divided into feature extraction stage $\boldsymbol{v} = f_{\theta} (\boldsymbol{x})$ and loss computation stage $\cL (\boldsymbol{v}, y)$. 
In this section, 
we present our novel  modifications for the current pipeline, to equip the model with ability of extracting global-scale and multiple local-scale features in the first stage 
and guide the model to evaluate the reliability and ID-relativity of each feature in the second stage. 

\subsection{Multi-scale Feature Learning} \label{sec:msfl}

Prevailing Person ReID methods use one-size-fits-all high-level embeddings from a deep convolutional network for all cases. 
To be specific, the coarse-resolution high-semantic embeddings from the last layer is leveraged to retrieve the images in the gallery database.
This might limit their accuracy on difficult examples caused by variance of  resolution, scale, pose and viewpoint, detector misalignment, and extremely similar identity as shown in Fig.~\ref{fig:challenge}. 
As analyzed in Sec.~\ref{sec:intro}, it is important to empower the model to extract features of all scales. 

To achieve this goal, ideally we would want to reason jointly across multiple scales of semantic abstraction. 
However simply concatenating high and low level embeddings suffers from severe performance degradation, which is verified by \cite{huang2017multi}. 
There are two reasons that naively combining low level features hurts the performance: 
\textbf{1).} Early layers lack of semantically abstracted features. In the neural network, 
the high-resolution features of low level (shape and color) usually lack representation power and 
high-semantic features of high level (objects and its parts) tend to lose information about the fine spatial details. 
The features of early layers lack representation power and thus 
attach to final features prematurely will likely yield unsatisfactory high error rates. 
\textbf{2).} The direct deep supervision on low level features altering the internal representation and making it premature. 
Modern Neural Networks abstract semantic concept level by level,  intermediate supervision influencing the early features to be optimized for the short-term and not for the final layers. This might improves the discriminative of shallow features, but collapses information required to generate high quality features in later layers. This effect becomes more pronounced when the first classifier is attached to an earlier layer as shown in \cite{huang2017multi}. 

\begin{figure*}
	\centering
	\includegraphics[width=0.8\linewidth]{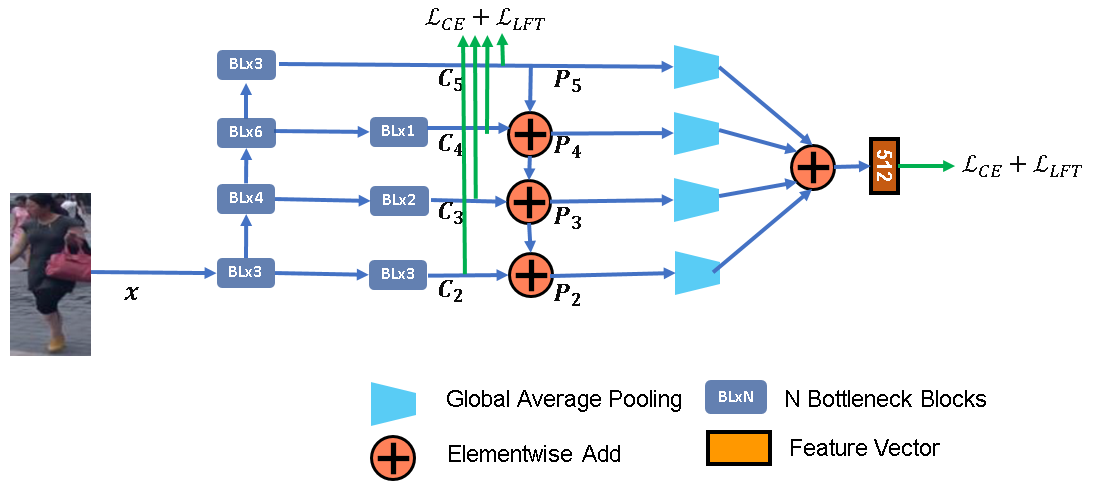}
	\caption{The overview of proposed methods.}
	\label{fig:net}
\end{figure*}

To remedy this issue, 
we present a novel Person ReID architecture that 
effectively solves the two problems by 
\textbf{1).} creating a  feature pyramid that has strong semantics at all scales, which naturally leverage the pyramidal shape of a CNN’s feature hierarchy. 
\textbf{2).} inserting more blocks according to the abstraction level of feature, 
postpone the abstraction of shallow feature and mitigate the effect of deep supervision. 
The overview of proposed network architecture is shown in Fig.~\ref{fig:net}.  
The architecture first extract features hierarchy and encode the image into a global context embedding via bottom-up path, 
then gradually propagate the information across different scales via top-down pathway and lateral connections, 
finally combine features of different scale via Multi-scale Feature Fusion (MSFF). 
There are two main consideration for the proposed network architecture: 

\textbf{Cross-scale information propagation (CSIP)} 
The clues to discriminate persons are imbalanced across the features of different scales and isolated from each other. 
To overcome these problem, we combine low and high level features and allow information communicate and rebalance across scales, via the design of top-down pathway and lateral connection. 

The first step is extracting features hierarchy. For the convenience of description, we define a \textit{stage} of a network as the combination of layers that produce feature maps of the same size. 
We choose the output of the last layer of each stage as our reference set of feature maps, which we will enrich to create the feature pyramid. This choice is natural since the deepest layer of each stage should have the strongest features.
Specifically, for ResNet backbone\cite{he2016deep}, we use the feature activations of each stage’s last residual block (conv1, conv2, conv3, conv4), denoted as $\{C_2, C_3, C_4, C_5\}$, with the strides of $\{4, 8, 16, 32\}$ \wrt the input image. 
We do not include conv1 into the pyramid due to its large memory footprint. 

Then, the top-down pathway upsamples the high-level strong-semantic features.  
These features are then enhanced with features from the bottom-up pathway via lateral connections. 
Each lateral connection merges feature maps of the same spatial size from the bottom-up pathway and the top-down pathway. The bottom-up feature map is of lower-level semantics, but its activations are of high resolution and more accurately localized. 

To name the details, we use nearest neighbor upsampling to  upsample the spatial resolution by a factor of 2, 
and append $3\times 3$ convolution to reduce the aliasing effect of upsampling.  
The lateral connection consists of a $1\times 1$ convolutional layer to normalize the channel dimensions to 512.
The outputs of lateral connection and upsapling module would be merged element-wise addition. 
This final set of feature maps is called $\{P_2, P_3, P_4, P_5\}$, corresponding to $\{C_2, C_3, C_4, C_5\}$ that are respectively of the same spatial sizes. 
  
The result is a feature pyramid that has rich semantics at all levels, where  the details of low level are conditionally decoded according to the global context propagated from high-semantic embeddings. 

\textbf{Multi-scale feature fusion (MSFF)}
Although low-level embeddings contain rich information about shape, color, and texture, they are not abstract enough to discriminate different persons.
Thus, we stack more bottleneck blocks into shallow layers, refine the feature representation, and finally integrate the information of different scales via concatenating. 
To specific, we further abstract features $\{P_2, P_3, P_4\}$ by 3, 2, and 1 bottleneck blocks respectively, the finally induced feature set is $\{F_2, F_3, F_4, P_5\}$. 

\subsection{Multi-scale Gradient Regularizer}

It is one thing that 
all clues across difference scale must be extracted and considered jointly by MSFL, 
while it is a more important thing that 
some clues should be permitted partially altered or even missing. 
To achieve the later and avoid increasing the computation burden in the test stage, 
we introduce an differentiable Multi-scale Gradient Regularizer (MSGR), 
which is applied in the training stage to help the model extract a more ID-related and discriminative embedding. 

Ross et al. \cite{ross2018improving} demonstrate that input gradient regularization helps the model be robust  to  adversarial attacks
while be more naturally interpretable. 
Inspired by \cite{ross2018improving}, we expect that MSGR applied to low-level feature maps could interpretably alter the ID-related color/context information, 
and MSGR applied to high-level feature maps could interpretably change the ID-related semantic part information. 
Altough the experiments show that the effect is more complicated and entangled, we describe the methods from simple to complex. 

We first describe single-scale Gradient Regularizer, which is naturally derived from the worst-case perturbation and intimatedly become a regularizer in the final loss function. 
Denote $\boldsymbol{x}$ as input images, and $\mTheta$ as model parameters. For simplicity, we denote  $\mathcal{L}(\boldsymbol{x}; \mTheta)$ as final loss, which encapsulate the feature extraction process. 
The idea could be formalized as follows. 
Instead of solving $\min_{\mTheta} \cL ( \boldsymbol{x}; {\mTheta})$, 
ideally we would like to solve the following problem 
if we try to build a robust model against any perturbation $\epsilon$. 
The perturbation is observed to be random at beginning stage and then become ID-related as the model converging. 

\begin{equation}
\label{eq:inner}
\min _{\boldsymbol{\theta}} \max _{\boldsymbol{\epsilon} :\|\boldsymbol{\epsilon}\|_{p} \leq \sigma} \mathcal{L}(\boldsymbol{x}+\boldsymbol{\epsilon} ; \boldsymbol{\theta})
\end{equation}

The norm constraint in eq.~\ref{eq:inner} implies that we only require our model to be robust against certain small perturbation (that do not change ID completely). 
In general, this problem is difficult to be solved explicitly due to its non-convex nature \wrt $\boldsymbol \epsilon$ and $\mTheta$. 
We propose to solve it via  first order Taylor expansion at point $\boldsymbol{x}$. 
The inner problem then becomes
\begin{equation}
\label{eq:taylor}
\max _{\boldsymbol \epsilon} \mathcal{L}(\boldsymbol{v})+\nabla_{\boldsymbol{v}} \mathcal{L}^{T} \boldsymbol{\epsilon} \quad \text { s.t. } \quad\|\boldsymbol{\epsilon}\|_{p} \leq \sigma
\end{equation}

This problem is trivially linear, and hence convex \wrt $\boldsymbol{\epsilon}$. We can obtain a closed form solution by Lagrangian multiplier method, see Appendix~\ref{sec:solving} for details. This yields 
\begin{equation}
\label{eq:epsilon}
\epsilon=\sigma \operatorname{sign}(\nabla \mathcal{L})\left(\frac{|\nabla \mathcal{L}|}{\|\nabla \mathcal{L}\|_{p^{*}}}\right)^{\frac{1}{p-1}}
\end{equation}
where $p^*$ is the dual of $p$, \ie, $\frac{1}{p^{*}}+\frac{1}{p}=1$. 

Substitute the optimal $\boldsymbol{\epsilon}$ back to the original optimization problem eq.~\ref{eq:inner}, we can see that the influence of perturbations can be formulated as a regularization term:
\begin{equation}
\min_{ \mTheta} \mathcal{L}(\boldsymbol{x}; \mTheta)
+\sigma\left\|\nabla_{\boldsymbol{x}} \mathcal{L}\right\|_{p^{*}} 
\label{eq:loss}
\end{equation} 

There are multiple ways to implement the regularization term: 
\textbf{1).} Imitate adversarial training methods, decompose the training procedure into two stages, first 
calculate the adversarial perturbation by Eq.~\ref{eq:epsilon},
and then feed the adversarial input and perform the ordinary training procedure to minimize the loss function by gradient descent on $\mTheta$.  
\textbf{2).} Summarize the influence of perturbations into a regularization term, as shown in Eq.~\ref{eq:loss}, and allow the gradient of $\left\|\nabla_{\boldsymbol{x}} \mathcal{L}\right\|_{p^{*}}$ \wrt $\mTheta$ flows back to $\mTheta$.  
Empirically, \textbf{2).} is superior than \textbf{1).}, which is also verified by adversarial attack and defense history, \ie, adversarial training methods usually harm the model performance and gradient regularizer helps  the model converge to a flatter local optimum. Thus, we directly optimize the regularization, 
leverage the auto-derivation ability provided by the modern deep learning framework. 
For case $p=\infty$,  regularization term $\|\nabla \cL\|_1$ is induced to penalized gradient in an anisotropic way,
as for $p=1$, the induced $\|\nabla \cL\|_\infty$ only penalizes the gradient in one direction. 
Empirically, experiments show that $p=2$ is more appropriate than both cases above.

Finally, as to multi-scale gradient regularizer, denote $\mTheta =\{\boldsymbol{\theta}_1, ... ,\boldsymbol{\theta}_K \}$ as model parameters; 
$\boldsymbol v = f_K( \cdots (f_1(\boldsymbol x; \boldsymbol{\theta}_K)); \boldsymbol{\theta}_1)$ as the features $\boldsymbol v$ that hierarchically extracted by $K$-stage neural network; 
and $\mathcal{L}(\boldsymbol{v} ; \mTheta)$ as a loss function.  
For MSFL that based on ResNet backbone, there are are $K=4$ scales of features. 
Instead of solving $\min_{{\mTheta}} \mathcal{L}(\boldsymbol{v} ; \mTheta)$, 
ideally we would like to solve the following problem if we try to build a robust model against small perturbation
$\{\boldsymbol \epsilon_1, ... ,\boldsymbol \epsilon_k \}$ 
on the features of any scale, including perturbation of texture, color and  semantically abstracted attributes such as pose, gender, race and clothing style. 
\begin{equation} 	
\label{eq:joint}
\min _{\mTheta} 
\max _{\boldsymbol{\epsilon}_k :\|\boldsymbol{\epsilon}_k\|_{p} \leq \sigma, \forall k}
\mathcal{L}(
f_K \left( \cdots (f_1(\boldsymbol x)+\boldsymbol \epsilon_1 ) )+\boldsymbol \epsilon_K 
; {\mTheta} \right) 
\end{equation}

Since the nested hierarchical function can be approximated by first order taylor expansion level by level, the inner problem is simplified as
\begin{equation}
\begin{aligned}
\max_{\forall i \in \{1,...,K\}, \|\boldsymbol{\epsilon}_i\|_{p} \leq \sigma}
 & \cL(f(\boldsymbol{x})) \\ 
 & + (\nabla_{f_K} \cL)^T \cdots (\nabla_{\boldsymbol{x}} f_1)^T \cdot \boldsymbol{\epsilon}_1  \\ 
 & + \cdots + \nabla_{\boldsymbol{f}_K} \cL^T  \boldsymbol{\epsilon}_K
\end{aligned}
\end{equation}

Applying chain rule, $ (\nabla_{f_K} \cL)^T \cdots (\nabla_{\boldsymbol{x}} f_1)^T \cdot \boldsymbol{\epsilon}_1$ is simplified as $\nabla_{\boldsymbol x} \cL^T \boldsymbol{\epsilon}_1$. Similarly, the loss with multi-scale gradient regularizer is 
\begin{equation}
\min_{ \mTheta} \mathcal{L}(\boldsymbol{x}; \mTheta)
+\sigma\left\|\nabla_{\boldsymbol{x}} \mathcal{L}\right\|_{p^{*}}  
+ \cdots 
+ \sigma\left\|\nabla_{{f_K}} \mathcal{L}\right\|_{p^{*}} 
\end{equation}

\section{Experiments} \label{sec:exp}

\subsection{Implementation Details} 

We follow the setting of widely used open-source implementation of 
 \cite{zhou2019omni}. 
To make sure fair comparison, we do not apply the tricks like Last Stride, Label Smooth \cite{szegedy2016rethinking} and LR Warmup, mentioned in \cite{luo2019bag}.
It is worthy to mention that although Luo \etal \cite{luo2019bag} achieves astonishing performance based on ResNet50 backbone \cite{he2016deep}, the dimension of feature embedding is 2048, and the Rank-1 accuracy on market1501 will drop from $94.5\%$ to $93.0\%$ if the dimension is reduced to $512$. 
During the training stage, our implementation details includes:

\textbf{1).} We initialize the ResNet50 with pretrained parameters on ImageNet. 
To reduce the feature dimension to $512$, which is the same as \cite{zhou2019omni}, we append BN-FC-ReLU layers after global average pooling of ResNet50, 
and change the output dimension of classifier to the number of identities in the training dataset. 

\textbf{2).} To constitute a training batch, $P$ identities and $K$ images of per person are sampled randomly. The batch size equals to $B = P \times K$. This approach has shown very good performance in similarity-based ranking and avoids the need to generate a combinatorial number of examplar pairs. 
During a training epoch each identity is selected in its batch in turn, and the remaining $P-1$ batch identities are sampled at random. We set $P=8$, and $K=4$. 

\textbf{3).} For image preprocessing and augmenting, we resize each image to $288 \times 144$ pixels,  randomly crop it into a $256 \times 128$ rectangular image, and flipped horizontally with 0.5 probability. Follow the fashion of ImageNet, each image is first divided by 255, normalized to $[0,1]$ range, substracted by $[0.485, 0.456, 0.406]$, and finally divided by $[0.229, 0.224, 0.225]$. These statistics are calculated according to the natural images in ImageNet. 

\textbf{4).} Adam is adopted to optimize the model. The initial learning rate is 0.00035 and decrease by a ratio of 0.1 at the 40th epoch and 70th epoch respectively. Totally there are 120 training epochs. We train the model two NVIDIA TITAN XP GPUs and Pytorch as the platform, with the float16 training  and  BatchNorm synchronized across GPU as default strategies. 


\subsection{Dataset and Protocol}

We focus on three widely-used large datasets: CUHK03\cite{li2014deepreid}, Market1501\cite{zheng2015scalable}, DukeMTMC-ReID\cite{zheng2017unlabeled}, and MSMT17\cite{wei2018person}. 

\textbf{CUHK03}\cite{li2014deepreid}dataset contains 13,164 images of 1,360 identities. It provides bounding boxes detected from deformable part models (DPMs) and manual labeling.
We adopt the new training/testing protocol raised by Zhong \etal\cite{zhong2017reranking} using a new training/testing protocol similar to that of Market-1501. The new protocol splits the dataset into training set and testing set, which consist of 767 identities and 700 identities respectively.
In testing, one image from each camera is randomly selected as the query for each identity and use the rest of images to construct the gallery set.
Compared to the traditional protocol that splits the dataset into  training set with 1,160 identities and  testing set with 100 identities,  
the new protocol has two advantages: 
\textbf{1).} For each identity, there are multiple ground truths in the gallery, which is more consistent with practical application scenario. 
\textbf{2).} Evenly dividing the dataset into training set and testing set at once helps avoid repeating training and testing multiple times. 

\textbf{Market1501}\cite{zheng2015scalable} dataset contains 32,668 images of 1,501 labeled
persons of six camera views. There are 751 identities in the training set and 750 identities in the testing set. 
In the original study on this proposed dataset, the author also uses mAP as the metric to evaluate the overall ranking quality of predicted rank list. 


\textbf{DukeMTMC-ReID}\cite{zheng2017unlabeled} is a subset of DukeMTMC\cite{ristani2016MTMC}
and contains 36,411 images of 1,812 identities captured by eight high-resolution cameras. 
The pedestrian images are cropped by hand-drawn bounding boxes. It consists of 16,522 training images of 702 identities, 2,228 query images and 17,661 gallery images of the other 702 identities. 

\textbf{MSMT17}\cite{wei2018person} is currently the largest Person ReID
dataset, which contains 126,441 images of 4,101 identities in 15 cameras. This dataset is composed of the training set, which contains 32,621 bounding boxes of 1,041 identities and the test set including 93,820 bounding boxes of 3,060 identities. From the test set, 11,659 images are used as query images and the other 82,161 bounding boxes are used as gallery images. This challenging dataset has more complex scenes and backgrounds, \eg, indoor and outdoor scenes, than others.

To evaluate the performance of proposed methods and compare with other ReID methods, we report two common evaluation metrics: the cumulative matching characteristics (CMC) at rank-1
and mean average precision (mAP) on the above four benchmarks following the common settings\cite{zheng2015scalable,wei2018person}.

\subsection{Visualization of Learned Features} 

To understand how MSFL+MSGR help learn discriminative features, we visualize the activation values of high-level feature maps $P_5$ to investigate which semantic parts the network focuses on. Following \cite{zhou2019omni}, the activation maps are computed as the sum of absolute-valued feature maps along the channel dimension followed by a spatial $L_2$ normalization. 

\begin{figure}
	\centering
	\includegraphics[width=0.99\linewidth]{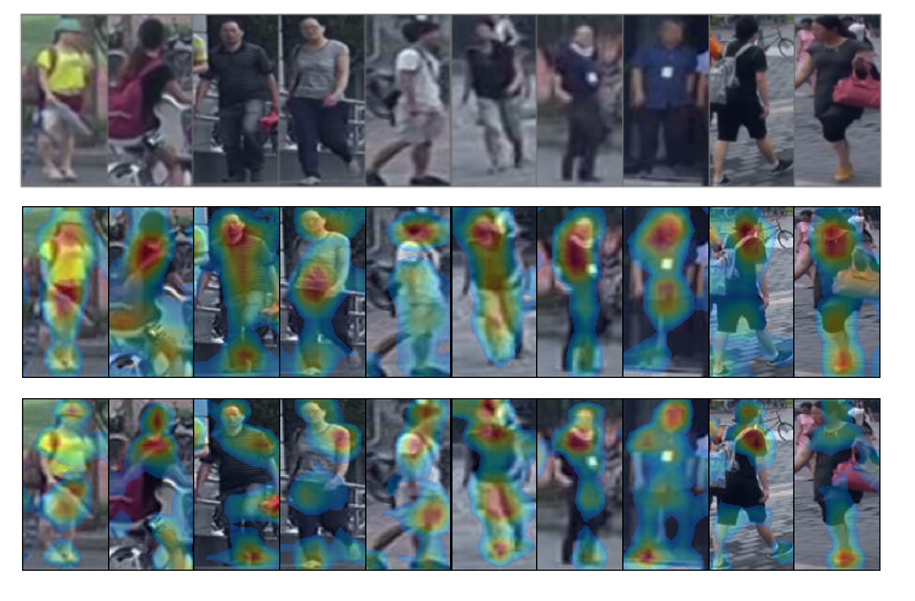}
	\caption{
		First line contains original images. Second line and third line contain the activation map of ResNet50 baseline and MSFL+MSGR respectively. 
		These images indicate that MSFL+MSGR detects subtle details like bags (Column 3), shoes (Column 9) and reliably clear frontal face (Column 4), to help discriminate visually similar persons.}
	\label{fig:activation}
\end{figure}

Fig. \ref{fig:activation} compares the activation maps of the ResNet50 baseline and
the model with MSFL+MSGR. 
It is clear that MSFL+MSGR can capture the local discriminative patterns of a person, such as clothing logo, bags (Column 3), shoes (Column 9) and reliably clear frontal face (Column 4), to distinguish person from person. 
In contrast, the baseline model over-concentrates on less informative region like background. 

\begin{figure}
	\centering
	\includegraphics[width=0.8\linewidth]{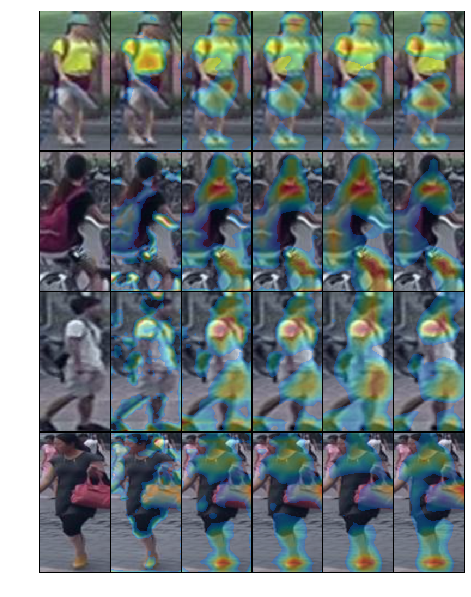}
	\caption{
		Each line contains a visual case. First Column is the original image; the second to the last column corresponds to activation maps of  $C_2,  P_2,  F_2, C_5, P_5$ of MSFL+MSGR respectively. 
		Due to lack of representative power, $C2$ may over-concentrate on parts or background with salient color (Row 1 Column 2 and Row 4 Column 2) and 
		edge and context without semantic meaning. With the help of global context from top-down path $P_2$ focus on discriminative person parts. 
		Meanwhile, $P_5$ attend to multiple discriminative parts which is helpful for robust ReID. 
	}
	\label{fig:unet}
\end{figure}

To understand how MSFL helps collect multi-scale discriminative clues, we visualize the feature $C_2,  P_2,  F_2, C_5, P_5$ in the hierarchy. As shown in Fig.~\ref{fig:unet}, due to lack of representative power, $C2$ may over-concentrate on parts or background with salient color (Row 1 Column 2 and Row 4 Column 2) and 
edge and context without semantic meaning. With the help of global context from top-down path $P_2$ focus on discriminative person parts. 
Meanwhile, $P_5$ attend to multiple discriminative parts which is helpful for robust ReID. 

To conclude, the qualitative results demonstrate that our multi-scale design, aggregation methods and regularization 
enable the model to identify subtle differences between visually similar persons –- a vital requirement for accurate ReID.

\subsection{Ablation Study}

We conduct ablation study on Market1501 dataset. 
Comparing (a) with (c) in Tab.~\ref{tab:msfl}, we verify that direct fusing the feature on deep stage can slightly improve the performance, however, directly incorporating the shallow feature (Tab.~\ref{tab:msfl}(b)) hurts the performance.
The reasons are discussed in Sec.~\ref{sec:msfl}, \ie, low-level features lack of semantic abstraction and direct deep supervision on low-level features altering the internal representation and making it premature. 

On the contrary, CSIP (Tab.~\ref{tab:msmt}(e)) allow the shallow feature to communicate with global context and be conditionally decoded, thus improve the rank-1 from 90.4\% to 91.5\%. 
CSIP consists of two complementary modules, \ie, top-down pathway and lateral connection. Lacking of top-down pathway hurts the performance  (Tab.~\ref{tab:msfl}(a) and (c)), 
and lacking of lateral connection cannot improve the performance (Tab.~\ref{tab:msfl}(a) and (d)). 

Meanwhile, MSFF (Tab.~\ref{tab:msmt}(f)) empower the shallow feature to abstract its detailed information with complex transformation and further improve the rank-1 to 92.3\%.  

\begin{table}
	\centering
	\caption{Ablation Study on Multi-Scale Feature Learning (MSFL)} 
	\label{tab:msfl}
	\scalebox{0.85}{ 
		\begin{tabular} {lccccc}
			\toprule 
			Methods	& lateral& top-down & MSFF & rank-1  & mAP  \\			\midrule						
			(a) ResNet50 &&& & 90.4   & 78.3  \\ \midrule 
			(b) direct fuse $C_4$ and $C_5$ & $\surd$ &&    & 90.7   & 78.6 \\ 
			(c) direct fuse from $C_2$ to $C_5$  & $\surd$ && & 87.1   & 75.9\\ \midrule 
			(d) fuse from $P_2$ to $P_5$, w.o. lateral  &  & $\surd$&  & 90.3   & 78.3 \\
			(e) fuse from $P_2$ to $P_5$, with lateral  & $\surd$ & $\surd$& & \textbf{91.5} & \textbf{79.8} \\
			(f) fuse from $F_2$ to $P_5$ & $\surd$ & $\surd$ &$\surd$& \textbf{92.3}   & \textbf{81.9}  \\			\bottomrule
		\end{tabular}
	} 
\end{table}	

Our design of MSFL modifies the network architecture with neglectable computation overhead. 
The Last stride=1 trick \cite{sun2018beyond} (Tab.~\ref{tab:complex}(b)) is widely used in Person ReID community which improves the flops from 2.6G to 4.1G. 
This trick removed the last spatial downsampling operation after stage 3 in the backbone network to increase the size of the feature map,  with the hope of improving performance by increasing spatial resolution. 
We do not need to apply this trick, since the MSFL maintains roughly equivalent amount of parameters and flops whiles become superior than Last stride=1 trick.

\begin{table}
	\centering
	\caption{Complexity of MSFL} 
	\label{tab:complex}
	\scalebox{0.85}{ 
		\begin{tabular} {lcccc}
			\toprule 
			Methods	& params (M)& flops (G) & rank-1  & mAP  \\			\midrule						
			(a) ResNet50 baseline & 24.56& 2.6& 90.4   & 78.3  \\ \midrule 
			(b) ResNet50, Last Stride=1 & 24.56 &4.1&  91.3   & 79.8 \\ 
			(c) fuse from $F_2$ to $P_5$  & 27 & 5.9 & \textbf{92.3}  & \textbf{81.9}\\ 
			\bottomrule
		\end{tabular}
	} 
\end{table}

As for MSGR, we first decide that $L_2$-norm is the most beneficial for regularizer term (Tab.~\ref{tab:MSGR}(a)--(d)), which meets our hypothesis that a model with smooth gradients of fewer extreme values is capable of extracting balanced ID-related information, and thus more robust and accurate. 
The improvement from Gradient Regularizer is stable (Tab.~\ref{tab:MSGR}(d)--(f)), which is not sensitive to the choice of $\sigma$ in a long range. 
Thus, we fix $\sigma = 1e-2$ in the latter experiments. 
Meanwhile, via regularizing the loss \wrt the output of middle stages to be smooth, 
the \textbf{Multi-scale} Gradient Regularizer further improve the performance. 
The most improvement happens when is applied to input and early stages, 
it may because the regularization on input gradient includes the effects to encourage the output of middle stages to be smooth, \ie, $\nabla_{\boldsymbol x} \cL^T  = (\nabla_{C_K} \cL)^T \cdots (\nabla_{\boldsymbol{x}} f_1)^T \cdot$.
However, it still improves the performance when applying the MSGR to middle outputs, since different stages is responsible for different levels of abstraction. In the latter experiments, we default adopt MSGR on the outputs of stage 1 and 2. 


\begin{table}
	\centering
	\caption{Ablation Study on Multi-scale Gradient Regularizer (MSGR)} 
	\label{tab:MSGR}   
	\scalebox{0.95}{ 
		\begin{tabular} {lcccc}
			\toprule 
			Methods & norm & $\sigma$ & rank-1 & mAP  \\			\midrule						
			(a) r50 baseline &/ &/ & 90.4  & 78.3  \\ \midrule  			
			(b) $\|\nabla_{\boldsymbol{x}} \cL \|$ & 1 & 1e-2 & 85.1 & 70.1  \\  
			(c) $\|\nabla_{\boldsymbol{x}} \cL \|$ & 2 & 1e-2 & \textbf{91.7} &	\textbf{80.1}  \\ 
			(d) $\|\nabla_{\boldsymbol{x}} \cL \|$ & $\infty$ & 1e-2 & 88.2 & 72.2 \\ 
            (e) $\|\nabla_{\boldsymbol{x}} \cL \|$ & 2 & 1e-1 & {90.7} &	79.4  \\ 		
			(f) $\|\nabla_{\boldsymbol{x}} \cL \|$ & 2 & 1e-3 & 91.2 &	{80.0}  \\ 	
			(g) $\|\nabla_{C_2} \cL \|$ & 2 &1e-2 & 92.1 &	80.4  \\
			(h) $\|\nabla_{C_5} \cL \|$ & 2 &1e-2 & 90.2  & 78.3  \\	\midrule 
			(i) $\|\nabla_{\boldsymbol{x}} \cL\| + \|\nabla_{C_2} \cL \|$ &2&1e-2&  92.5 &	\textbf{80.3} \\ 
			(j) $\|\nabla_{\boldsymbol{x}} \cL\| + \sum_{k=2}^{3} \|\nabla_{C_k} \cL \|$ &2&1e-2&  \textbf{92.7} &	80.1\\ 
			(k) $\|\nabla_{\boldsymbol{x}} \cL\| + \sum_{k=2}^{4} \|\nabla_{C_k} \cL \|$ &2&1e-2&  {92.5} &	80.1 \\ 
			(l) $\|\nabla_{\boldsymbol{x}} \cL\| + \sum_{k=2}^{5} \|\nabla_{C_k} \cL \|$ &2&1e-2&   {92.4} & 79.8 \\  
			\bottomrule
		\end{tabular}
	}				
\end{table}

%

\subsection{Comparison with the state-of-the-art methods}

\begin{table*}[htbp]
	\centering
	\caption{Comparison of the state-of-the-art results on four large benckmarks. 
		The CMC scores (\%) at rank-1 and mAP are listed. 
		To show the effectiveness of each component, we gradually stack them to the ResNet50 baseline.}
	\scalebox{1.15}{ 
	\begin{tabular}{lrrrrrrrr}
		\toprule
		& \multicolumn{2}{c}{Market1501} & \multicolumn{2}{c}{Detected CUHK03} & \multicolumn{2}{c}{DukeReID} & \multicolumn{2}{c}{MSMT17} \\
		& \multicolumn{1}{l}{Rank-1} & \multicolumn{1}{l}{mAP} & \multicolumn{1}{l}{Rank-1} & \multicolumn{1}{l}{mAP} & \multicolumn{1}{l}{Rank-1} & \multicolumn{1}{l}{mAP} & \multicolumn{1}{l}{Rank-1} & \multicolumn{1}{l}{mAP} \\ \midrule
		PAN   & 82.2  & 63.3  & 36.3  & 34    & 71.6  & 51.5  &       &  \\
		SVDNet \cite{sun2017svdnet}  & 82.3  & 62.1  & 57.1  & 54.2  & 76.7  & 56.8  &       &  \\
		PDC  \cite{su2017pose} & 84.1  & 63.4  &       &       &       &       & 58    & 29.7 \\
		HAP2S \cite{yu2018hard} & 84.6  & 69.4  &       &       & 75.9  & 60.6  &       &  \\
		DPFL \cite{chen2017person}  & 88.6  & 72.6  & 40.7  & 37    & 79.2  & 60.6  &       &  \\		
		DaRe\cite{huang2017multi}  & 86.4  & 69.3  & 55.1  & 51.3  & 74.5  & 56.3  &       &  \\
		DaRe+RE & 89    & 76    & 63.3  & 59    & 80.2  & 64.5  &       &  \\
		PNGAN \cite{qian2018pose} & 89.4  & 72.6  &       &       & 73.6  & 53.2  &       &  \\
		GLAD \cite{wei2017glad} & 89.9  & 73.9  &       &       &       &       & 61.4  & 34 \\
		KPM  \cite{shen2018end} & 90.1  & 75.3  &       &       & 80.3  & 63.2  &       &  \\
		MLFN \cite{chang2018multi} & 90    & 74.3  & 52.8  & 47.8  & 81    & 62.8  &       &  \\
		DuATM \cite{si2018dual} & 91.4  & 76.6  &       &       & 81.8  & 64.6  &       &  \\
		Bilinear \cite{suh2018part} & 91.7  & 79.6  &       &       & 84.4  & 69.3  &       &  \\
		G2G \cite{shen2018deep}  & 92.7  & 82.5  &       &       & 80.7  & 66.4  &       &  \\
		DeepCRF \cite{chen2018group}  & 93.5  & 81.6  &       &       & 84.9  & 69.5  &       &  \\
		PCB+RPP \cite{sun2018beyond} & 93.8  & 81.6  & 63.7  & 57.5  & 83.3  & 69.2  &       &  \\
		SGGNN  \cite{shen2018person} & 92.3  & 82.8  &       &       & 81.1  & 68.2  &       &  \\
		Auto-ReID+RE \cite{quan2019autoreid} & \textbf{95.4} & \textbf{94.2} & \textbf{73.3} & \textbf{69.3} & \textbf{91.4} & \textbf{89.2} & \textbf{78.2} & \textbf{52.5} \\ 
		Mancs \cite{wang2018mancs} & 93.1  & 82.3  & 65.5  & 60.5  & 84.9  & 71.8  &       &  \\
		OSNet \cite{zhou2019omni} & 93.6  & 81    & 57.1  & 54.2  & 84.7  & 68.6  & 71    & 43.5 \\
		OSNet+RE & 94.8  & 84.9  & 72.3  & 67.8  & 88.6  & 73.5  & 78.7  & 52.9 \\ \midrule
		ResNet50 & 90.4  & 78.3  & 63.4  & 58.3  & 82.9  & 70.6  & 6.7  & 38.9 \\
		+CSIP & 91.5  & 79.3  & 65.6  & 60.4  & 84.1  & 74.5  & 69.3  & 40.2 \\
		+MSFF & 92.3  & 81.9  & 65.1  & 61.2  & 85.7  & 75.4  & 70.1  & 40.8 \\
		+MSGR & \textbf{93.7} & \textbf{83.6} & \textbf{67.7} & \textbf{63.3} & \textbf{86.8} & \textbf{76.9} & \textbf{71.3} & \textbf{45.3} \\
		+RE   & 94.4  & 89.5  & 71.3  & 68.6  & 89.6  & 89    & 78.4  & 51.7 \\ \bottomrule
	\end{tabular}%
	}
	\label{tab:mktabl} \label{tab:cuhk03} \label{tab:msmt} \label{tab:duke} 
\end{table*}%

On Market101 dataset (Tab.~\ref{tab:mktabl}, via gradually stacking each component to the ResNet50 model, 
we improve the rank-1 from 90.4\% to 93.7\%, outperform the most of the state-of-the-art methods, and reach 94.4\% when applying Rerank. 
It worth to mention thath we build our model upon strong baseline, and improve the performance steadily without exhausted hyper-parameter tuning. 


On CUHK03 dataset (Tab.~\ref{tab:cuhk03}), due to misalignment, the performance on detected CUHK03 is worse than labeled CUHK03. 
Compared to Market1501, the improvement brought by MSFF is relatively small, but by further applying MSGR, the rank-1 on labeled CUHK03 improves to 71.6\%. 
We presume that the training data scarcity of new CUHK03 protocol causes the optimization difficulty of MSFF block, 
but MSGR is an appropriate regularizer that helps smooth the loss landscape and reduces the optimization difficulty.

On DukeMTMC-ReID (Tab.~\ref{tab:duke}), the effect of each method is similar to that on Market1501. 
MSFL+MSGR achieves rank-1 of 86.8\%, superior than the most state-of-the-art model, 
and even as competitive as Auto-ReID that applies NAS to automatic discover the best architecture for ReID task. 
On MSMT17 (Tab.~\ref{tab:msmt}), despite of complex scenes and diverse backgrounds, MSFL+MSGR improves rank-1 metric from 82.9\% to 86.8\%.

\section{Conclusion}
Person ReID faces many challenges, which are summarized as two aspects, 
\ie, the dramatic scale variance and unreliability of each clue for distinguish person. 
To address these problems, this paper propose 
a model with potential to discover global-scale and multiple local-scale clues
and 
introduce adversarial learning to encourage the model to mine subtle clues while determine ID-relativity of each clue by multi-scale gradient regularizer. 

It worth to mention that the architecture and regularizer designed by our experiences may not be the optimal.
In our future work, we may apply  the idea of AutoML to discover more superior model architecture and loss function that tailored for Person ReID task.

\appendices

\section*{Acknowledgment}
This work was partially supported by 

\section{Solving $\epsilon$ in Eq.~\ref{eq:taylor}} \label{sec:solving}

We need to solve $\epsilon$ from Eq.~\ref{eq:taylor}, which  is equivalent to 
\begin{equation}
\max _{\epsilon} \nabla_{\boldsymbol{x}} \mathcal{L}^{T} \boldsymbol{\epsilon} \quad \text { s.t. } \quad\|\boldsymbol{\epsilon}\|_{p} \leq \sigma
\end{equation}

The optimal $\epsilon$ would be achieved when $\|\boldsymbol{\epsilon}\|_{p}=\sigma$, otherwise, we can increase the norm of $\epsilon$ and increase the objective value. Thus, we are set to solve
\begin{equation}
\max _{\epsilon} \nabla_{\boldsymbol{x}} \mathcal{L}^{T} \boldsymbol{\epsilon} \quad \text { s.t. } \quad\|\boldsymbol{\epsilon}\|_{p}=\sigma
\end{equation}

By introduce Lagrangian multiplier $\lambda$, we have 
\begin{align} 
\nabla_\epsilon \nabla_{\boldsymbol{x}} \mathcal{L}^{T} \boldsymbol{\epsilon} &=\lambda \nabla_\epsilon \| \epsilon \|_p  \\ 
\nabla_{\boldsymbol{x}} \mathcal{L} &=\lambda \frac{\boldsymbol{\epsilon}^{p-1}}{p\left(\sum_{i} \boldsymbol{\epsilon}_{i}^{p}\right)^{1-\frac{1}{p}}}  \\ 
\nabla_{\boldsymbol{x}} \mathcal{L} &=\frac{\lambda}{p}\left(\frac{\boldsymbol{\epsilon}}{\sigma}\right)^{p-1} \label{eq:comb1} \\
\nabla_{\boldsymbol{x}} \mathcal{L}^{\frac{p}{p-1}} &=\left(\frac{\lambda}{p}\right)^{\frac{p}{p-1}}\left(\frac{\boldsymbol{\epsilon}}{\sigma}\right)^{p} 
\end{align}

Sum each element of the vector over two sides, we have 
\begin{align}
\|\nabla \mathcal{L}\|_{p^{*}}^{p^{*}} &=\left(\frac{\lambda}{p}\right)^{p^{*}} * 1 \\
\left(\frac{\lambda}{p}\right) &=\|\nabla \mathcal{L}\|_{p^{*}} \label{eq:comb2}
\end{align}

Combine Eq.~\ref{eq:comb1} and Eq.~\ref{eq:comb2}, it is easy to see
\begin{equation}
\epsilon=\sigma \operatorname{sign}(\nabla \mathcal{L})\left(\frac{|\nabla \mathcal{L}|}{\|\nabla \mathcal{L}\|_{p^{*}}}\right)^{\frac{1}{p-1}}
\end{equation}

{	
	\bibliographystyle{ieee}
	\bibliography{egbib}
}

\end{document}